\documentclass[lettersize,journal]{IEEEtran}
\usepackage[ruled,vlined]{algorithm2e}
\usepackage{amsmath,amsfonts}
\usepackage{array}
\usepackage[caption=false,font=normalsize,labelfont=sf,textfont=sf]{subfig}
\usepackage{textcomp}
\usepackage{stfloats}
\usepackage{url}
\usepackage{verbatim}
\usepackage{colortbl}
\usepackage{tcolorbox}
\usepackage{graphicx}
\usepackage{cite}
\begin{document}

\title{Fitness Landscape of Large Language Model-Assisted Automated Algorithm Search}


\author{Fei Liu, Qingfu Zhang, Jialong Shi, Xialiang Tong, Kun Mao and Mingxuan Yuan
\IEEEcompsocitemizethanks{\IEEEcompsocthanksitem Fei Liu and Qingfu Zhang are with the Department of Computer Science, City University of Hong Kong, Hong Kong (e-mail: fliu36-c@my.cityu.edu.hk; qingfu.zhang@cityu.edu.hk). Jialong Shi is with School of Mathematics and Statistics, Xi’ an Jiaotong University, China. Xialiang Tong and Mingxuan Yuan are with HUAWEI Noah's Ark Lab. Kun Mao is with Huawei Cloud EI Service Product Department.
}
}
\maketitle

\begin{abstract}
Using Large Language Models (LLMs) in an evolutionary or other iterative search framework have demonstrated significant potential in automated algorithm design. However, the underlying fitness landscape, which is critical for understanding its search behavior, remains underexplored. In this paper, we illustrate and analyze the fitness landscape of LLM-assisted Algorithm Search (LAS) using a graph-based approach, where nodes represent algorithms and edges denote transitions between them. We conduct extensive evaluations across six algorithm design tasks and six commonly-used LLMs. Our findings reveal that LAS landscapes are highly multimodal and rugged, particularly in combinatorial optimization tasks, with distinct structural variations across tasks and LLMs. Moreover, we adopt four different methods for algorithm similarity measurement and study their correlations to algorithm performance and operator behaviour. These insights not only deepen our understanding of LAS landscapes but also provide practical insights for designing more effective LAS methods.

\end{abstract}





\section{Introduction}

Large Language Models (LLMs) assisted evolutionary or iterative search methods have been used for automated algorithm design with great success~\cite{liu2024evolution}~\cite{yao2025multi}~\cite{romera2024mathematical}~\cite{ye2024reevo}~\cite{van2024llamea}~\cite{van2024loop}~\cite{wu2024evolutionary}. These methods cast an algorithm design task as an optimization problem and try to find an optimal algorithm in an algorithm space in an iterative manner. Leveraging LLM capabilities in language comprehension, reasoning and code generation, they are able to conduct effective and efficient search in an algorithm space~\cite{liu2024systematic}. However, there is lack of good understanding on why and when these methods work or not. Traditional math analysis can be hardly used to study these methods due to the complexity of the algorithm space~\cite{zhang2024understanding} and the black-box nature of LLMs~\cite{allen2023physics}.

Fitness landscape analysis~\cite{malan2013survey,zou2022survey,ochoa2014local,shi2020homotopic,huang2025fitness,rodrigues2022fitness,de2024landscape,wang2024effects,thomson2025stalling} is widely used for investigating behaviors of search methods, particularly, in the area of heuristics. Very recently, \cite{van2025code} has made a first attempt to study the landscape of LLM-assisted algorithm design methods. They introduced Abstract Syntax Tree (AST)-based Code Evolution Graphs (CEGs) and investigated the relationship between candidate algorithms~\footnote{For the sake of clarity, we call an algorithm in the algorithm space a candidate algorithm} (i.e., an element in the algorithm space) and their characters in feature space. However, CEGs focus on the structural properties of the algorithms found during the search, while fitness landscape analysis is less discussed.

In this paper, we focus on the fitness landscape of LLM-assisted Algorithm Search (LAS). Leveraging a graph-based approach, we visualize and analyze the fitness landscape, providing deeper insights into the search dynamics of LAS. Specifically, we try to answer the following research questions:

\begin{itemize}
\vspace{5pt}
\item \textbf{Q1:} What does the fitness landscape of LLM-driven algorithm search look like, and what features (e.g., plateaus, ruggedness) can be used to characterize it?
\vspace{2pt}
\item \textbf{Q2:} How does the landscape vary across different algorithm design tasks, LLMs, and settings? 
\vspace{2pt}
\item \textbf{Q3:} How can we measure the similarity in algorithm space, and what are the correlations of these similarity measurements and algorithm behaviour?
\vspace{5pt}
\end{itemize}

Our major contributions are:
\begin{itemize}
    \item We cast LLM-assisted Algorithm Search (LAS) as an optimization problem over the algorithm space, where each candidate algorithm is represented as code (code implementation) and/or thought (algorithm idea description) as in ~\cite{liu2024evolution,van2024llamea}. 
    \item We use a graph-based approach to visualize and analyze the fitness landscape to enhance our understanding of its structure. Each node represents one candidate algorithm generated in the search process, and the edges are the connections between algorithms during the search. 
    \item We conduct a comprehensive evaluation across six distinct algorithm design tasks and six different LLMs. We find the fitness landscape of LLM-assisted algorithm search is usually multi-modal with distinct patterns and characteristics across different algorithm design tasks and LLMs.
    \item  We evaluate four code-based similarity metrics for measuring distances between algorithms in code representation. While we demonstrate correlations between code similarities and performance, validating that different evolutionary operators provide varying degrees of exploration in the algorithm space, our results also reveal high variability in these distance measurements, suggesting the need for new similarity metrics specifically tailored for LAS.
\end{itemize}

\section{Formulation of LLM-assisted Algorithm Search}

We formulate the LLM-assisted automatic Algorithm Search (LAS) as follows. Consider a target algorithm design task $T$ (e.g., design a heuristic for the Traveling Salesman Problem (TSP)) with a set of instances $I$, 
The goal is to find an optimal algorithm $a^*$  in an algorithm space $A^{S}$ that minimizes a performance metric $F(a, I)$, which is defined as the average of performance scores of $a$ across all instances $i \in I$. Without loss of generality, in this paper, we consider the minimization problem. Mathematically, it can be expressed as:

\begin{equation}
    F(a, I) = \frac{1}{|I|}\sum_{i \in I} f(a, i), \quad a \in A^{S}.
\end{equation}

The optimal algorithm $a^*$ is obtained by solving the following optimization problem:
\begin{equation}
a^* = \underset{a \in A^{S}}{\arg\min} \, F(a, I).
\end{equation}

The optimization problem in the algorithm space $A^S$ can be solved  by  evolutionary search~\cite{liu2024systematic,zhang2024understanding}, where each algorithm is represented as code and/or text. Starting from an initial population of algorithms $P_0$, the evolutionary search iteratively conducts three key steps: generation, evaluation, and update, until a stopping criterion is met. It works as follows: 

\textbf{Step 1: Initialization} The initial population of candidate algorithms $P_0$ is either produced by prompting LLMs or selected from 
existing hand-crafted algorithms.

\textbf{Step 2: Iterative Process} To produce $P_{t+1}$ from $P_t$, it does the following:
\begin{itemize}
    \item \textbf{Step 2.1: Generation} LLMs are prompted using instructions and existing parent candidates $a_p$, which are selected from the current population $P_i$. This can be represented as 
    $$a_o = LLM(Prompt(a_p)),
    $$ 
    where $a_p$ and  $a_o$ denote the parent and offspring candidates, respectively. It should be pointed out that $a_p$ and/or $a_o$ can be multiple candidate algorithms. 
    $LLM$ represents the LLM inference process with instructions. $Prompt()$ is the instruction used for LLMs.

    \item \textbf{Step 2.2: Evaluation} The fitness $F(a, I)$ of $a$ is obtained  by evaluating $a$ on the instance set $I$.

    \item \textbf{Step 2.3: Update} Add some selected candidates from  $a_o$ to $P_t$ and remove some candidates from $P_t$ to generate $P_{t+1}$.
    
\end{itemize}
\textbf{Step 3: Termination} The search process terminates when a stopping criterion is met. This criterion could be a maximum number of iterations, a maximum number of LLM requests, a performance threshold, or other problem-specific conditions. Upon termination, the best-performing algorithm $\hat{a}$, along with any other relevant information, will be the output.






\begin{algorithm}[htb]
\SetAlgoLined
\KwIn{Instance set $I$, initial population generation method, selection strategy, stopping criterion}
\KwOut{Best algorithm $\hat{a}$ found}

\BlankLine
\textbf{Step 1: Initialization} \\
Generate initial population $P_0$ either by prompting LLMs or selecting from existing algorithms;\\
Evaluate initial population;\\
$t \gets 0$

\BlankLine
\textbf{Step 2: Iterative Process} \\
\While{stopping criterion not met}{
    \textbf{Generation:} Select parent(s) $a_p \in P_t$ according to selection strategy;\\
    Generate offspring $a_o \gets \text{LLM}(\text{Prompt}(a_p))$;\\
    \textbf{Evaluation:} Compute fitness $F(a_o, I) \gets \frac{1}{|I|}\sum_{i \in I} f(a_o, i)$;\\
    \textbf{Update:} $P_{t+1} \gets \text{UpdatePopulation}(P_t, a_o)$;\\
    $t \gets t + 1$;
}

\BlankLine
\textbf{Step 3: Termination} \\
\Return{$\hat{a} \gets \underset{a \in  P_k}{\arg\min} \, F(a, I)$.}
\caption{LLM-assisted Algorithm Search (LAS)}
\end{algorithm}

\begin{figure}[htbp]
    \centering
    \includegraphics[width=\linewidth]{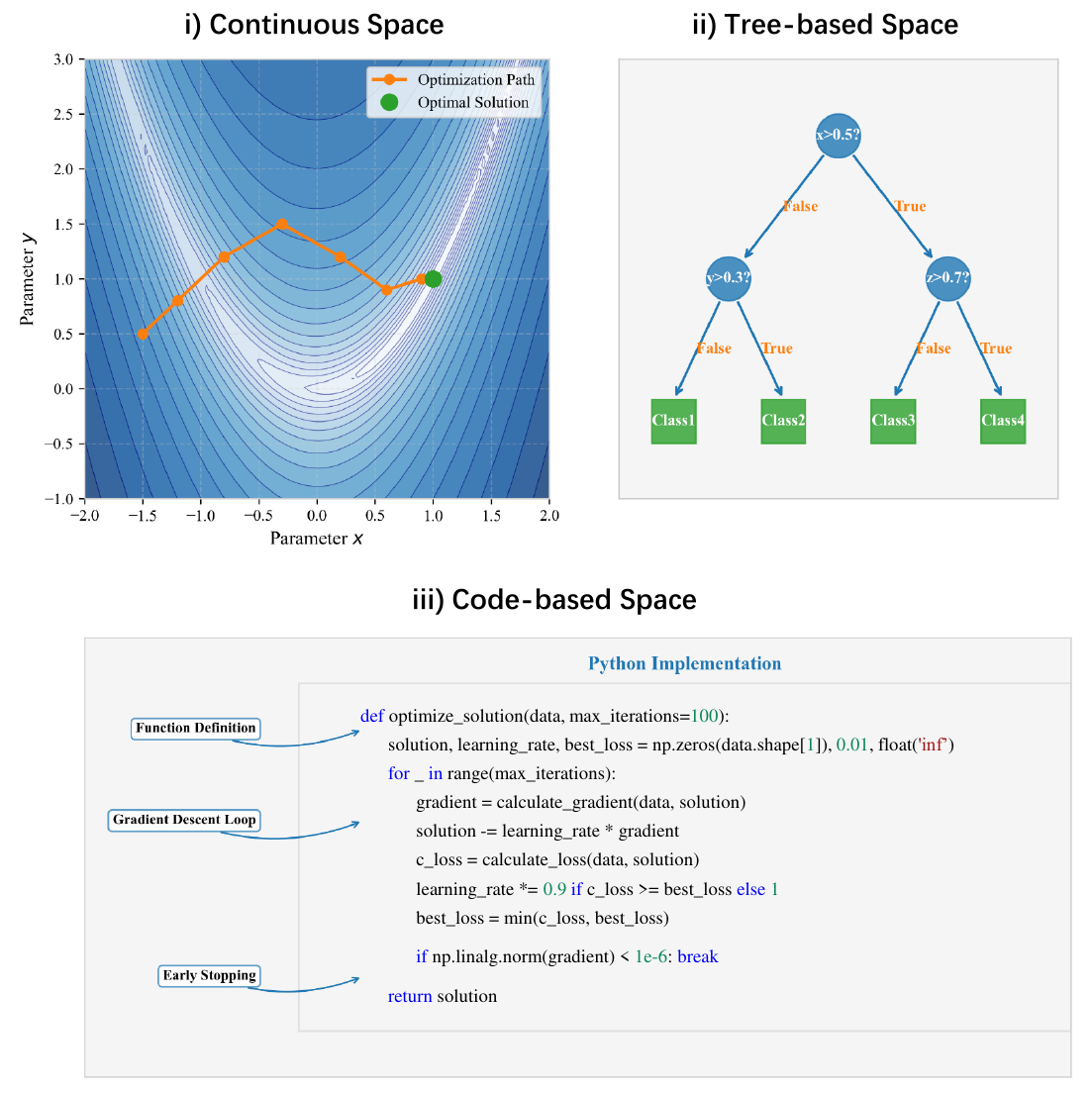}
    \caption{Three distinct algorithm representations are commonly used in automated algorithm search methods. \textbf{i)} For \textbf{continuous/discrete parameter repesentation}, many existing optimization techniques can effectively handle the optimization of fitness landscapes. \textbf{ii)} In the case of \textbf{tree-based/structured representations}, methods such as genetic programming are often employed. \textbf{iii)} \textbf{Code-based/text-based representations}, on the other hand, offer greater flexibility and power. Recent advancements in LLMs have enabled direct operation and search within this complex space. However, it poses significant challenges in understanding the corresponding fitness landscape. }
    \label{fig:representation}
\end{figure}

\subsection{Algorithm Representation}

Prior to the advent of LLMs, most existing works~\cite{burke2013hyper,stutzle2018automated,pillay2021automated,ma2025toward} on automated algorithm design can be classified into two categories based on their approaches for defining the algorithm search space $a \in A^S$: 
\begin{itemize}
    \item[(i)] \textbf{Continuous/discrete space:} $A^S$ is parameterized as a (continuous, discrete, or mixed) parameter space~\cite{kerschke2019automated, pushak2022automl}, which can be solved as a classic optimization problem.
    \item[(ii)] \textbf{Tree-based/structure-based space:} Each algorithm $a$ is expressed as a tree, graph, or other structures~\cite{meng2021automated}, and search methods such as genetic programming can be used to find a better algorithm~\cite{langdon2013foundations,mei2022explainable,zhang2023survey}.
\end{itemize}
These representations have been extensively studied. Many well-established optimization techniques can be used.

Unlike traditional approaches, LAS methods usually represent each candidate algorithm $a$ using a code implementation~\cite{hemberg2024evolving} and/or linguistic description~\cite{liu2024evolution}. This representation is more expressive than conventional methods. However, because this space is not rigidly structured, it is challenging to understand the landscape and navigate it effectively. Fig.~\ref{fig:representation} contrasts three distinct algorithm representations: (1) regular continuous/discrete space, (2) tree-based/structure-based space, and (3) code-based/text-based space.

\subsection{Algorithm Generation}
In the context of LLM-assisted algorithm search, the generation of an algorithm can be expressed as:

$$
a_o = \text{LLM}(\text{Prompt}(a_p)),
$$
where the offspring algorithms $a_o$ are produced by prompting the LLM with specific instructions $\text{Prompt}()$, using parent algorithms $a_p$ as input.  In a single-point iterative search, one offspring is generated from a single parent in each iteration. In contrast, population-based search maintains and evolves a population of candidate algorithms, where both the number of parent and offspring algorithms typically exceeds one. Both the prompt strategies employed and the capabilities of the LLMs will influence the algorithm generation and thus may lead to different landscapes and performance.

\subsection{Evaluation}

The evaluation $F(a, I)$ maps each algorithm $a$ to an indicator of its performance (e.g., the average route length on TSP instances). The evaluation is conducted on a set of representative instances $I$. Without loss of generality, in this paper, we consider the minimization problem.

\section{Graph-Based Representation of Fitness Landscape}

\begin{figure*}
    \centering
    \includegraphics[width=\textwidth]{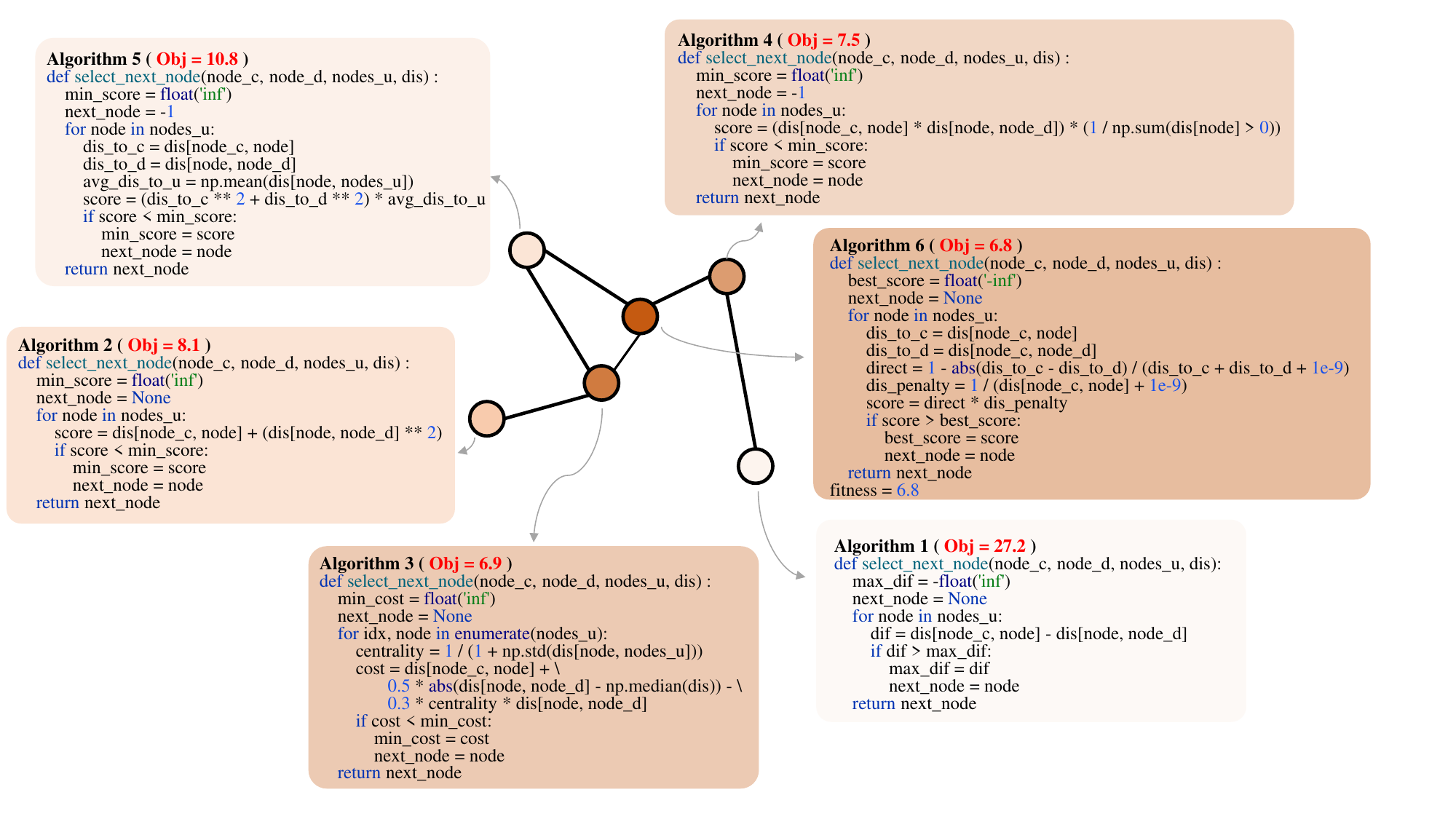}
    \caption{An illustration of a two-dimensional graph representation of fitness landscape, where nodes and edges are algorithms and their connections, respectively. The color of each node indicates the objective value (the smaller the better). Each algorithm is a constructive heuristic for the traveling salesman problem. The constructive heuristic determines how to select the next node to iteratively construct a solution. The inputs "node\_c", "node\_d", "nodes\_u" and "dis" are the current node ID, destination node ID, unvisited node IDs and distance matrix, respectively. The output is the next node ID.}
    \label{fig:illustration}
\end{figure*}

This section introduces a graph-based approach to studying the fitness landscape of LLM-assisted algorithm search. 
We begin by defining the concept of fitness landscapes and then elaborate on their graph-based representation.

A fitness landscape, as defined by~\cite{stadler2002fitness}, can be represented using a triplet $(S, NS, f)$, where:
\begin{itemize}
    \item $S$ is the set of potential solutions;
    \item $NS$ is the neighborhood structure, a function that assigns to each solution $s \in S$ a set of neighbors $N(s)$;
    \item $f: S \rightarrow \mathbb{R}$ is the objective function (fitness function), which quantifies the quality of each solution.
\end{itemize}


In the context of LLM-assisted algorithm search, the solutions $S$ are candidate algorithms generated by LLMs. The objective function $f$ is defined by specific performance metrics relevant to the task at hand. Since their are no well-defined distance metrics or neighbourhood structure~\cite{ochoa2008study,ochoa2014local} in the algorithm space, the neighbourhood is instead characterized by correlations between algorithms. Similar to Search Trajectory Networks (STNs)~\cite{ochoa2020search} and Attractor Networks~\cite{thomson2025stalling}, where no local optima are required and nodes can be any search space location, we adapt these concepts to LLM-assisted algorithm search. We define the neighborhood of an algorithm $a_i$ relative to another algorithm $a_j$ as follows: $a_i \in NS(a_j)$ if $a_i$ is generated from $a_j$ (i.e., $a_i = LLM(Prompt(a_j))$).  This formulation enables us to develop a graph-based representation that effectively illustrates the fitness landscape.

The graph-based representation is defined as follows:
\begin{itemize}
    \item \textbf{Nodes: } Each node in the graph represents an algorithm $a_i$ generated during the search process. The set of all nodes in the graph is denoted by $N$, with its size represented as $n$. The size of each node represents the number of this algorithm generated during the search process (the same algorithm can be generated many times during the search process).
    \item \textbf{Edges: } Edges in the graph are undirected and represent the genetic relationship between a parent algorithm and its offspring. Specifically, an edge exists from algorithm $a_i$ to algorithm $a_j$ if $a_j$ is directly derived from $a_i$ through the application of LLM-based search operators, i.e., $a_j = LLM(Prompt(a_i))$. Edges are weighted based on the frequency of the transition, where the weight corresponds to the number of times the transition (i.e., the generation of $a_j$ from $a_i$) occurs during the search process.
    \item \textbf{Graph: } The fitness landscape is represented as a graph $G = (N, E)$, where $N$ is the set of nodes corresponding to algorithms, and $E$ is the set of edges denoting parent-offspring relationships. We consider an undirected graph, where there is no direction between $a_i$ and $a_j$ if they are connected.
\end{itemize}

Fig.~\ref{fig:illustration} illustrates a two-dimensional graph representation of the fitness landscape for the traveling salesman problem. In this visualization, each node represents a distinct constructive heuristic algorithm, while edges indicate the connections between these algorithms during the LAS search process. The colour intensity of each node corresponds to its objective value, with darker shades indicating better performance (lower values). Each algorithm shown is a constructive heuristic that determines the next node selection strategy when iteratively building a TSP solution. These algorithms take several inputs: "node c" (the current node ID), "node d" (the destination node ID), "nodes u" (the set of unvisited node IDs), and "dis" (the distance matrix). Based on these inputs, each algorithm outputs the next node ID to visit, effectively defining different strategies for traversing the graph to construct a complete tour.

\section{Landscape of LLM-assisted Automated Algorithm Search}

\subsection{Experimental Settings}
We conduct Large Language Model-based Evolutionary Search (LES) on six algorithm design tasks using six widely used open-source and commercial LLMs. Our experiments employ EoH~\cite{liu2024evolution,liu2023algorithm}, a standard LLM-assisted evolutionary search framework for algorithm discovery. EoH follows a simple evolutionary structure, making it suitable for analyzing fitness landscapes.

We initialize the search using EoH's default prompt to generate initial algorithms and apply its four LLM-based evolutionary operators (E1, E2, M1, and M2) to explore new candidate algorithms. Each run maintains a population size of 20 and evaluates up to 2,000 candidate algorithms, which is the typical settings for LES~\cite{liu2024evolution,ye2024reevo,van2024llamea}.

\paragraph{Tasks}
We evaluate LES across six tasks spanning multiple domains:

\begin{itemize}
\item \textbf{Combinatorial Optimization}:
\begin{itemize}
\item \textbf{Online Bin Packing (OBP)}~\cite{romera2024mathematical}: Design a heuristic algorithm that assigns incoming items to bins in real time to minimize the total number of bins used.
\item \textbf{Traveling Salesman Problem (TSP)}~\cite{liu2023algorithm}: Design a constructive heuristic algorithm that selects the next node given a partial route and unvisited nodes to minimize the total route length.
\item \textbf{Capacitated Vehicle Routing Problem (CVRP)}~\cite{liu2024llm4ad}: Design a constructive heuristic algorithm that selects the next node while satisfying capacity constraints, minimizing the total route length.
\item \textbf{Vehicle Routing Problem with Time Windows (VRPTW)}~\cite{liu2024llm4ad}: Design a constructive heuristic algorithm that selects the next node while satisfying both capacity and time window constraints, minimizing the total route length.
\end{itemize}
\item \textbf{Agent Design}:
\begin{itemize}
\item \textbf{Mountain Car Task (Car)}~\cite{liu2024llm4ad}: A reinforcement learning task from OpenAI Gym~\cite{brockman2016openai} where the goal is to optimize the car’s actions to reach a target position with minimal iterations, adhering to position and velocity constraints.
\end{itemize}
\item \textbf{Symbolic Regression}:
\begin{itemize}
\item \textbf{Bacterial Growth Curve Approximation (Bact)}~\cite{shojaee2025llmsr}: A biology-inspired task aiming to model bacterial growth patterns by minimizing the mean squared error with respect to environmental parameters. The inclusion of symbolic regression allows us to examine algorithm/function design in a simpler, equation-based setting.
\end{itemize}
\end{itemize}


\paragraph{LLMs}
We test six LLMs of varying scales and capabilities:
\begin{itemize}
    \item \textbf{OpenAI}: GPT-3.5-Turbo, GPT-4o-Mini
    \item \textbf{DeepSeek}: DeepSeek-V3, DeepSeek-Chat
    \item \textbf{Other Open Models}: Qwen2-Turbo, GLM-3-Turbo
\end{itemize}

All experiments are executed on the open-source LLM4AD Platform~\cite{liu2024llm4ad}. We record all generated algorithms and their parent-offspring relationships during evolution, constructing a fitness landscape graph for each run.

\subsection{Landscape on Different Tasks}
\begin{figure*}
\centering
\includegraphics[width=1\linewidth]{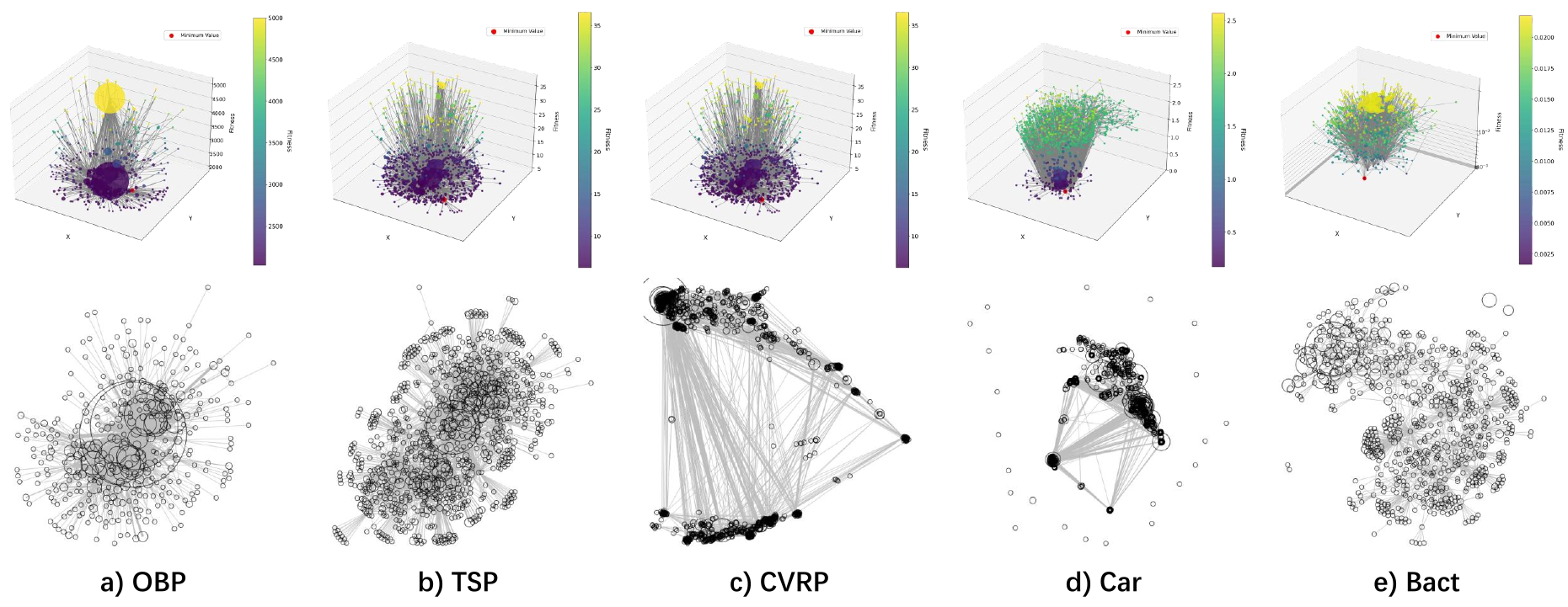}
\caption{The 2D and 3D fitness landscapes for five algorithm design tasks: Online Bin Packing (OBP), Traveling Salesman Problem (TSP), Capacitated Vehicle Routing Problem (CVRP), Mountain Car (Car), and Bacterial Growth (Bact).}
\label{fig:tasks}
\end{figure*}

\begin{figure}
\centering
\includegraphics[width=\linewidth]{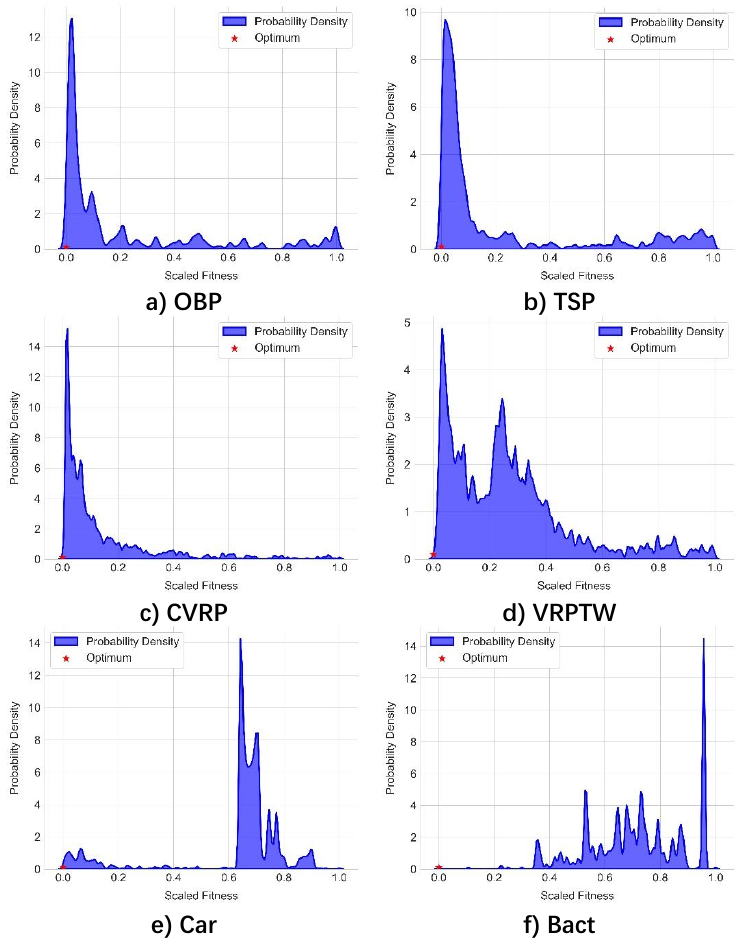}
\caption{The scaled probability distribution of fitness values across six tasks.}
\label{fig:distribution}
\end{figure}

In this section, we illustrate and analyze the fitness landscapes across different algorithm design tasks. Fig.~\ref{fig:tasks} illustrates the 2D and 3D visualizations of the fitness landscapes for five distinct tasks. In the 3D visualizations, the z-axis represents fitness values, where smaller (darker) values indicate better performance. The red node denotes the algorithm with the best value. Fig.~\ref{fig:distribution} shows the probability distribution of algorithm fitness values, which are normalized to the range [0, 1].

Based on the results, we make the following observations:
\begin{itemize}
\item The fitness landscapes exhibit several characteristics: \textbf{1) Multi-modal:} The search landscape exhibits a multi-modal nature, characterized by the presence of multiple approximated local optima. This feature highlights the complexity of the optimization problem, as the search process must navigate through numerous potential solutions rather than converging to a single global optimum. \textbf{2) Absence of Funnel Structure}: The landscape lacks a distinct funnel shape, which is often observed in many combinatorial optimization problems. This absence can be attributed to the inherent randomness in LLM-driven heuristic generation and the lack of formulated restrictions or guiding principles. As a result, the search process does not naturally converge toward a single optimal region but instead explores diverse areas of the solution space. \textbf{3) Low Connectivity}: The graph representation of the search landscape demonstrates low connectivity, as evidenced by the low average degree and sparsity of connections. This feature suggests that the search process operates in a fragmented manner, with limited interactions between nodes (solutions). Such low connectivity may impact the efficiency of information exchange and the propagation of beneficial traits across the population.
\item The fitness landscapes vary significantly across tasks. In heuristic design tasks, many algorithms are concentrated in the low-fitness region. This is evident in the distribution plots, where approximately 80\% of the algorithms fall within the lowest 20\% of fitness values. The VRPTW exhibits a slightly different distribution, likely due to the additional complexity introduced by time window constraints, making it more challenging to reach elite (low-fitness) algorithms.
\item The Car task displays a bimodal distribution, with most algorithms having poor fitness values and a small subset reaching the lowest end. Notably, there is a conspicuous absence of algorithms in the middle range of the fitness distribution.
\item In contrast to the other tasks, the optimal algorithm is exceptionally difficult to achieve in symbolic regression tasks. 98\% of the algorithms exhibit significantly higher objective values and are far from the optimal. This can be attributed to the nature of symbolic regression, where even minor mutations in functions or their parameters often lead to substantial changes in performance~\cite{zheng2025cst,shojaee2025llmsr}.
\end{itemize}

\subsection{Landscape using Different LLMs}

\begin{figure}[htbp]
    \centering
    \includegraphics[width=1\linewidth]{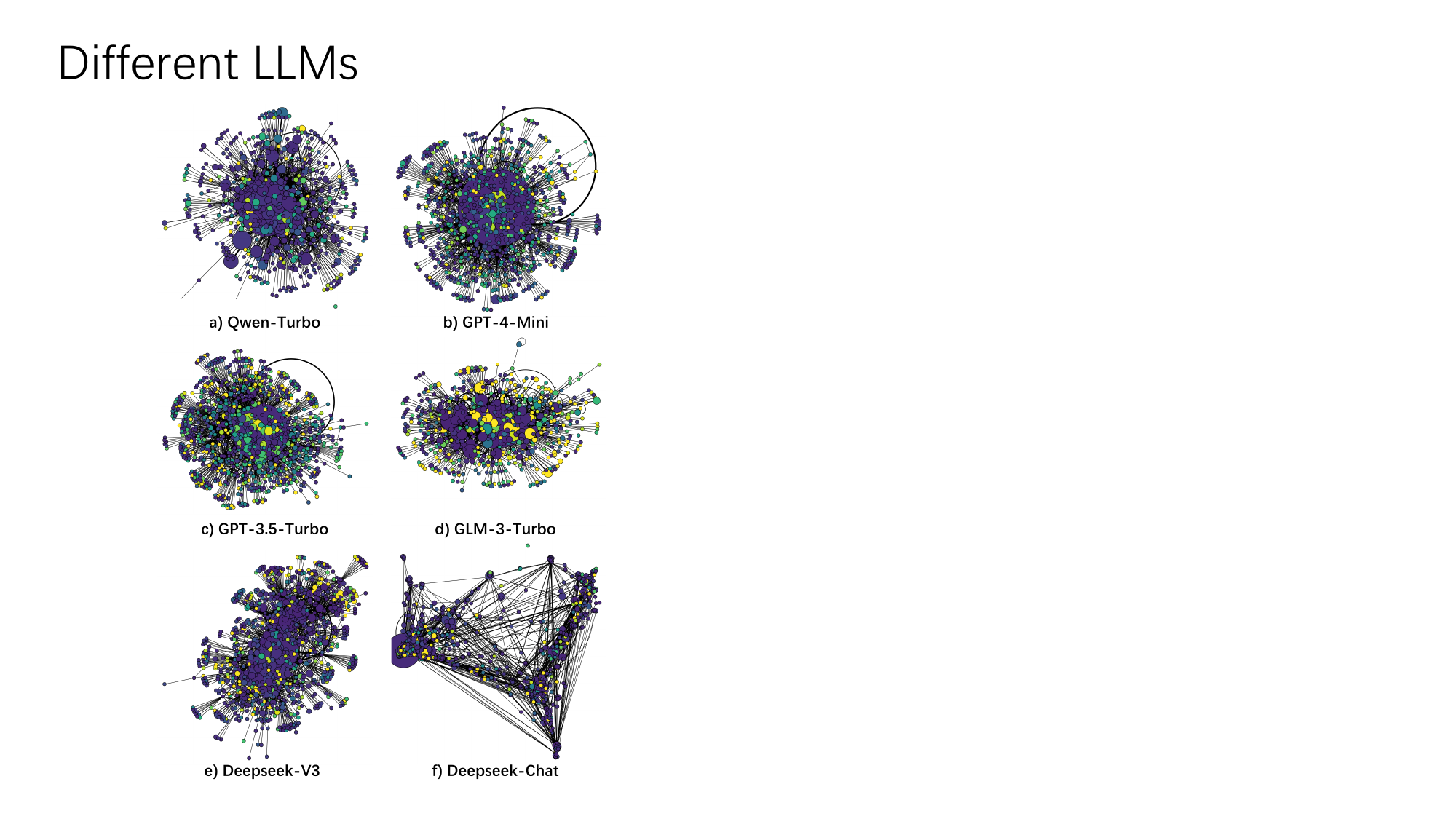}
    \caption{The 2D fitness landscapes of TSP using different LLMs}
    \label{fig:llms}
\end{figure}

In this section, we discuss the landscapes generated by different LLMs when applied to TSP task. Three metrics were used to evaluate graph characteristics: \textbf{Density}, the ratio of actual to possible edges (0-1), with higher values indicating greater connectivity; \textbf{Average Degree}, the mean number of edges per node, reflecting overall network interconnectedness; and \textbf{Clustering Coefficient}, which measures the tendency of nodes to form clusters (0-1), indicating the presence of local community structures. The results are summarized in Table~\ref{table:metrics}. Additionally, the visualizations of these landscapes are illustrated in Fig.~\ref{fig:llms}.

The metrics in Table~\ref{table:metrics} reveal variability in the structural properties of the solution landscapes generated by different LLMs. GLM-3-Turbo demonstrated the highest graph density (0.00223) and clustering coefficient (0.02282), indicating a highly interconnected and locally-focused exploration of the solution space. In contrast, DeepSeek-Chat produced the sparsest landscape (density = 0.00069) with the lowest average degree (2.15053) and clustering coefficient (0.00180), reflecting a more targeted and globally-oriented search strategy that minimizes redundant transitions. Qwen-Turbo, with the highest average degree (4.76190), exhibited broader exploration of neighboring solutions, while other models like GPT-3.5-Turbo and GPT-4o-Mini displayed intermediate behaviours, balancing local refinement and global exploration. These differences highlight distinct trade-offs in exploration strategies across the LLMs.

The graph visualizations in Fig.~\ref{fig:llms} reveal structural differences in the solution landscapes generated by the LLMs. Qwen-Turbo and GLM-3-Turbo produce densely connected, highly clustered landscapes, suggesting intensive local exploration that may improve refinement but risk local optima entrapment. GPT-3.5-Turbo and GPT-4o-Mini exhibit moderate density, balancing local and global exploration for adaptability. DeepSeek-V3 shows a sparser, elongated structure, prioritizing global exploration over local refinement. DeepSeek-Chat’s landscape is the sparsest and least clustered, indicating a targeted global search strategy that avoids redundancy and focuses on promising regions.

These results highlight a trade-off: models like GLM-3-Turbo and Qwen-Turbo excel in local refinement but may struggle with local optima, while DeepSeek-Chat’s sparse connectivity aids global exploration at the cost of local refinement. GPT-3.5-Turbo and GPT-4o-Mini strike a balance, enhancing robustness across diverse problems. Practically, dense solution landscapes are well-suited for tasks requiring intensive local optimization, while sparse landscapes facilitate broader exploration. Combining these behaviors could enhance the search process, as ensembles of diverse LLMs, as also noted in recent works~\cite{novikov2025alphaevolve}, often lead to better LES results.

\begin{table}[htbp]
\caption{Metrics using Different LLMs on TSP}
\label{table:metrics}
\centering
\begin{tabular}{lllll}
\hline
LLM           & Density                                  & Degree                          & Clustering Coef.                 \\ \hline
GLM-3-Turbo     & {\color[HTML]{6200C9} \textbf{0.00223}} & 4.04754                                 & {\color[HTML]{6200C9} \textbf{0.02282}} \\
Qwen-Turbo     & 0.00213                                & {\color[HTML]{6200C9} \textbf{4.76190}} & 0.00769                                 \\
GPT-3.5-Turbo  & 0.00110                               & 2.79492                                 & 0.00409                                 \\
GPT-4o-Mini    & 0.00135                             & 3.63397                                 & 0.00626                                 \\
DeepSeek-V3   & 0.00114                                & 2.91158                                 & 0.00533                                 \\
DeepSeek-Chat   & {\color[HTML]{FE0000} \textbf{0.00069}} & {\color[HTML]{FE0000} \textbf{2.15053}} & {\color[HTML]{FE0000} \textbf{0.00180}} \\ \hline
\end{tabular}
\end{table}

\subsection{Landscape using Different Population Sizes}

\begin{figure*}
    \centering
    \includegraphics[width=1\linewidth]{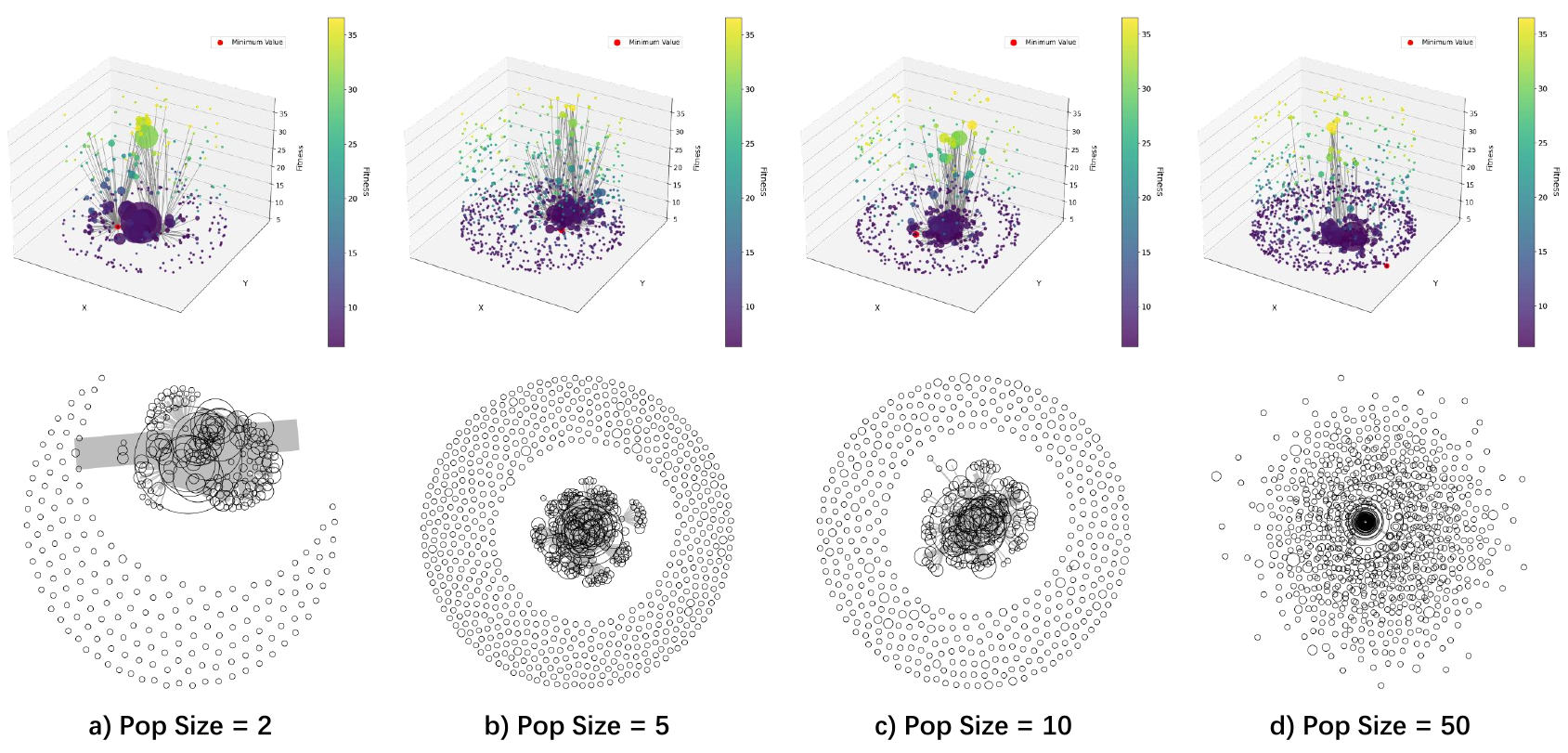}
    \caption{Visualization of landscapes with varying population sizes using the DeepSeek-V3 model on the TSP task. Each population size is depicted through both 3D and 2D representations, where candidate nodes denote distinct candidates within the population. To emphasize significant interactions, only edges with weights greater than 1 are displayed, indicating strong connections and frequent transactions between the connected nodes.}
    \label{fig:pop_size}
\end{figure*}

In this section, we present the experimental results of landscape exploration using the DeepSeek-V3 model applied to the TSP task. The visualizations in Fig.~\ref{fig:pop_size} provide insights into the structural dynamics of the search space as the population size varies. Both 3D and 2D representations are used to illustrate the distribution of candidates (nodes) and their interactions (edges), with edges weighted above 1 highlighted in orange to emphasize significant connections. Overall, smaller population sizes (e.g., 2 and 5) focus on local exploitation due to limited exploration, while larger sizes (e.g., 10 and 50) enable broader exploration, uncovering diverse solutions and revealing the landscape's inherent complexity with multiple unexplored peaks. 


\subsection{Evolution Trajectory}
\begin{figure}
    \centering
    \includegraphics[width=1\linewidth]{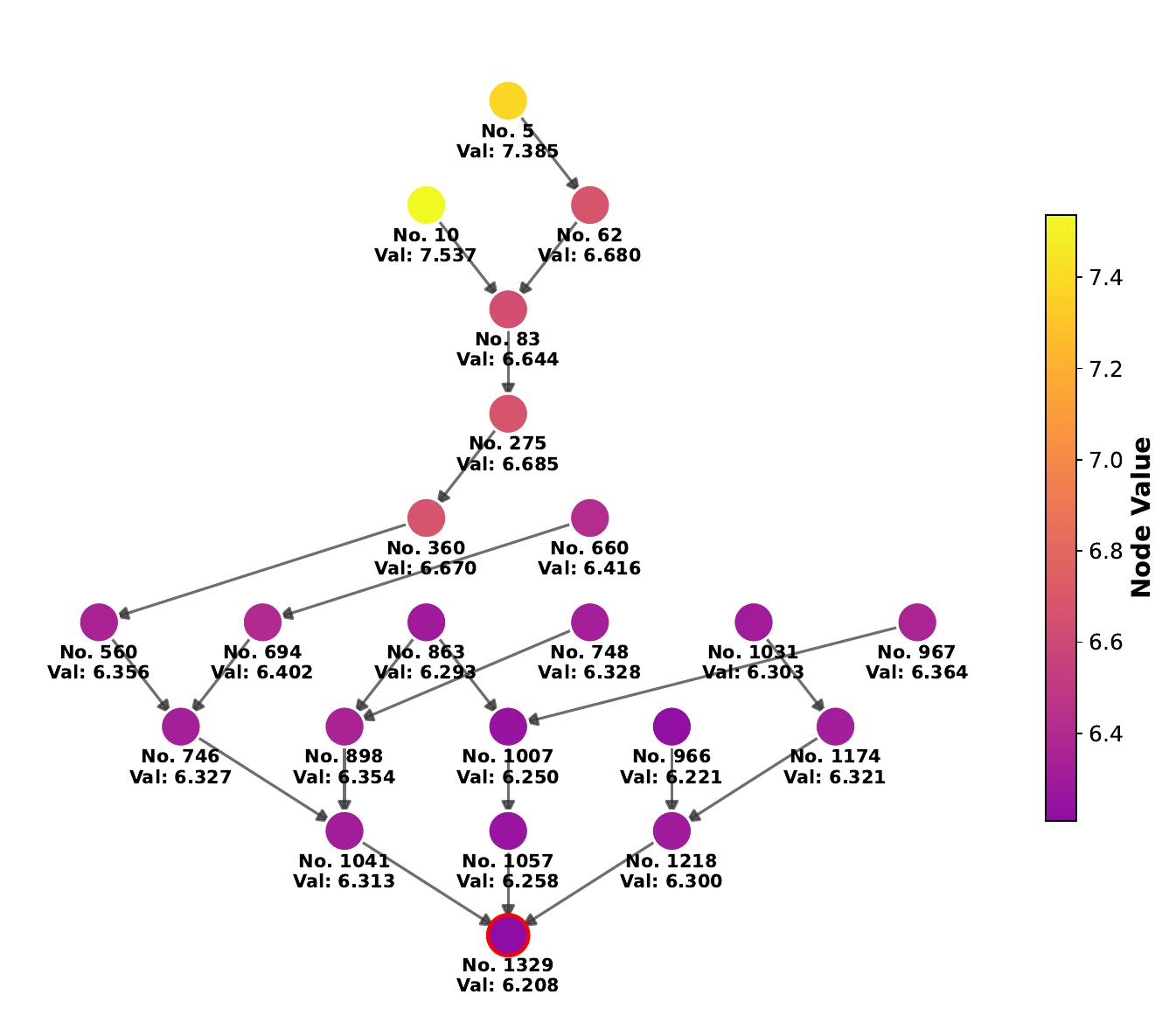}
    \caption{A directed network of candidate algorithms evolving toward the optimal algorithm (highlighted in red), where node colors represent fitness values and edges denote transitions between solutions. Nodes are arranged hierarchically, with labels showing node ids (i.e., the order during LLM-assisted search) and fitness values.}
    \label{fig:trajectory}
\end{figure}

Fig.~\ref{fig:trajectory} illustrates the evolutionary trajectory of algorithms discovered through LLM-based search using DeepSeek-V3 on TSP. Each node in the graph represents a distinct algorithm, with node values indicating performance according to our evaluation metric (i.e., the average route length on a set of TSP instances). We use directed edges to show the trajectory. The edge is direct from parent algorithm to offspring algorithm. The graph is structured hierarchically, with vertical positioning corresponding to the evolutionary path distance from the optimal algorithm (highlighted with a red circle).

This visualization captures several key aspects of our search process: (1) the branching exploration patterns as the LES investigated multiple promising directions, (2) the convergence toward increasingly effective algorithms, and (3) the critical evolutionary pathways that led to significant performance improvements. The network structure reveals both direct improvements and instances where seemingly suboptimal intermediate algorithms served as essential stepping stones toward superior algorithms. By tracing the edges from earlier algorithms to the optimal solution, we can identify the sequence of transformations that yielded the greatest performance gains, offering insights into the most productive modification strategies for this problem domain.

\subsection{Algorithm Similarity}

While edges on graph represent connections between algorithms (nodes), the lengths of edges do not associate with the actual distances. In this section, we study the similarity between algorithms and their correlation to performance and search operators. Because the algorithms are not hard-coded, there are no tailored techniques for directly measuring the distance or similarity between them. However, as we represent each algorithm as code, we can borrow ideas from code similarity measure techniques. In this section, we investigate four different code-based similarity measurements:

\begin{figure*}[htbp]
    \centering
    \includegraphics[width=\textwidth]{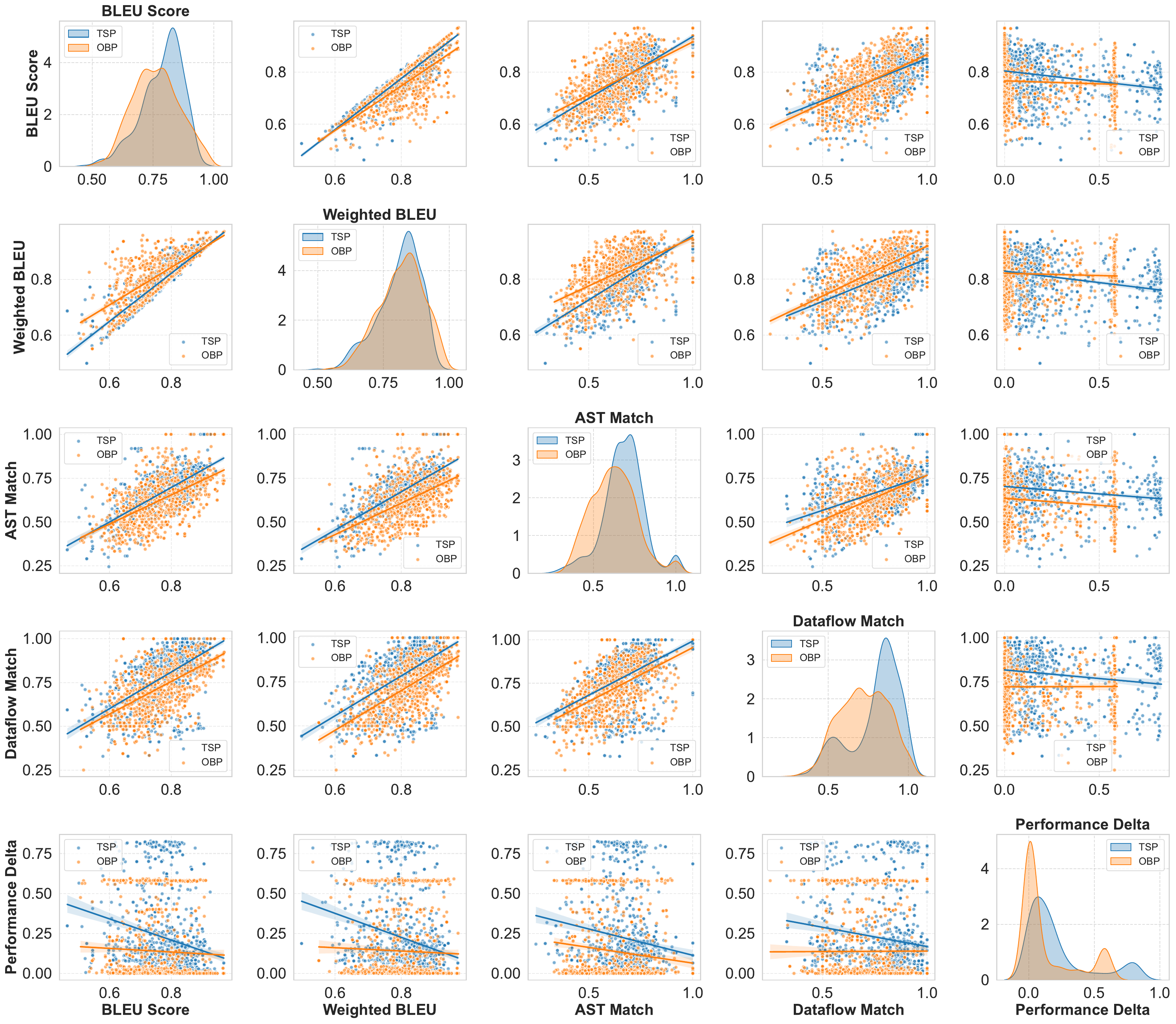}
    \caption{Matrix of metric relationships across TSP and OBP Tasks. It presents a 5x5 matrix of subplots visualizing the relationships between five key metrics: BLEU Score, Weighted BLEU, AST Match, Dataflow Match, and Performance Delta, across TSP and OBP. Each diagonal subplot represents the distribution of a single metric for both tasks, using Kernel Density Estimation (KDE) plots with shaded areas indicating the density of data points. These plots are color-coded by task, with blue for TSP and orange for OBP. These off-diagonal subplots show the relationships between pairs of metrics for each task. Scatter plots with regression lines are used to illustrate how one metric correlates with another. The correlation coefficient for each task is also displayed, providing insight into the strength and direction of these relationships.}
    \label{fig:pairplot}
\end{figure*}

\begin{itemize}
    \item \textbf{BLEU}~\cite{papineni2002bleu}: Computes the overlap of n-grams between two code snippets, evaluating lexical similarity. Originally designed for natural language translation evaluation, BLEU calculates precision scores for n-grams (typically n=1-4) and applies a brevity penalty for shorter outputs. Standard BLEU treats all tokens equally, which may not reflect the importance of keywords in programming languages and often fails to capture structural or semantic equivalence when code is refactored while maintaining identical functionality.
    
    \item \textbf{Weighted BLEU}~\cite{ren2020codebleu}: Enhances BLEU by assigning higher weights to programming keywords (e.g., \texttt{int}, \texttt{return}, \texttt{for}, \texttt{while}) in unigram matches. This prioritizes critical syntactic elements, improving relevance for code evaluation. By emphasizing language-specific constructs, Weighted BLEU better distinguishes between superficial token changes and modifications to core algorithmic structures, though it still primarily captures surface-level similarities rather than deeper semantic equivalence.
    
    \item \textbf{AST Match}~\cite{ren2020codebleu}: Measures syntactic similarity by comparing subtrees in the Abstract Syntax Tree (AST) of candidate and reference code. Leaf nodes (e.g., variable names) are ignored to focus on structural correctness. This approach captures program structure independently of variable naming conventions or formatting differences, making it more robust to cosmetic code variations. AST Match calculates the percentage of matching subtrees between two ASTs, providing insight into structural similarity that n-gram methods cannot detect.
    
    \item \textbf{Data-flow Match}~\cite{guo2020graphcodebert,ren2020codebleu}: Evaluates semantic similarity by analyzing variable dependencies in data-flow graphs. This metric constructs graphs representing how data values propagate through the program, with nodes as variables and edges as dependencies. Mismatches in value propagation (e.g., incorrect return variables, different computational paths) are detected, capturing functional discrepancies that might be missed by purely syntactic measures. Data-Flow Match is particularly effective at identifying semantically equivalent implementations with different syntactic structures, though its computation is usually more complex and resource-intensive than the other metrics.
\end{itemize}


\vspace{0.5em}
\noindent
To assess the effectiveness and inter-relation of these metrics, we compare them across TSP and OBP algorithm design tasks. Fig.~\ref{fig:pairplot} visualizes these comparisons in a 5×5 matrix of subplots covering five key metrics: BLEU Score, Weighted BLEU, AST Match, Dataflow Match, and Performance Delta (the relative performance difference between two algorithms). The diagonal subplots show the distribution of each metric for both tasks using kernel density estimation (KDE), with shaded density curves color-coded by task (blue for TSP, orange for OBP). Off-diagonal subplots display pairwise relationships between metrics, using scatter plots with fitted regression lines to highlight potential trends. For each task, the corresponding correlation coefficient is annotated, indicating the strength and direction of the relationship.

\paragraph{Inter-metric Correlation.}
As shown in the matrix plot (Fig.~\ref{fig:pairplot}), BLEU, Weighted BLEU, AST Match, and Dataflow Match exhibit strong positive correlations with each other in both TSP and OBP tasks. Notably, Dataflow Match and AST Match yield the highest correlations with performance delta for TSP, suggesting their alignment with functional correctness. 

\paragraph{Distributional Differences.}
Density plots along the diagonal of Fig.~\ref{fig:pairplot} reveal clear differences in score distributions between tasks. TSP examples are generally skewed toward higher AST and Dataflow Match scores, implying that effective solutions for TSP tend to preserve structural and semantic code features. In contrast, OBP examples display more dispersed scores, likely due to greater implementation diversity.

\paragraph{Performance Predictability.}
The bottom row of Fig.~\ref{fig:pairplot} evaluates how well each metric predicts performance delta. For TSP, all similarity metrics exhibit an inverse trend—higher similarity corresponds to lower performance delta—indicating that small modifications in similar code often lead to minor behavioral changes. Conversely, OBP shows weaker or inconsistent trends, especially for BLEU and AST Match, suggesting that similarity metrics alone maybe insufficient to predict performance variations in certain domains.

\begin{figure}[tbp]
    \centering
    \includegraphics[width=0.45\textwidth]{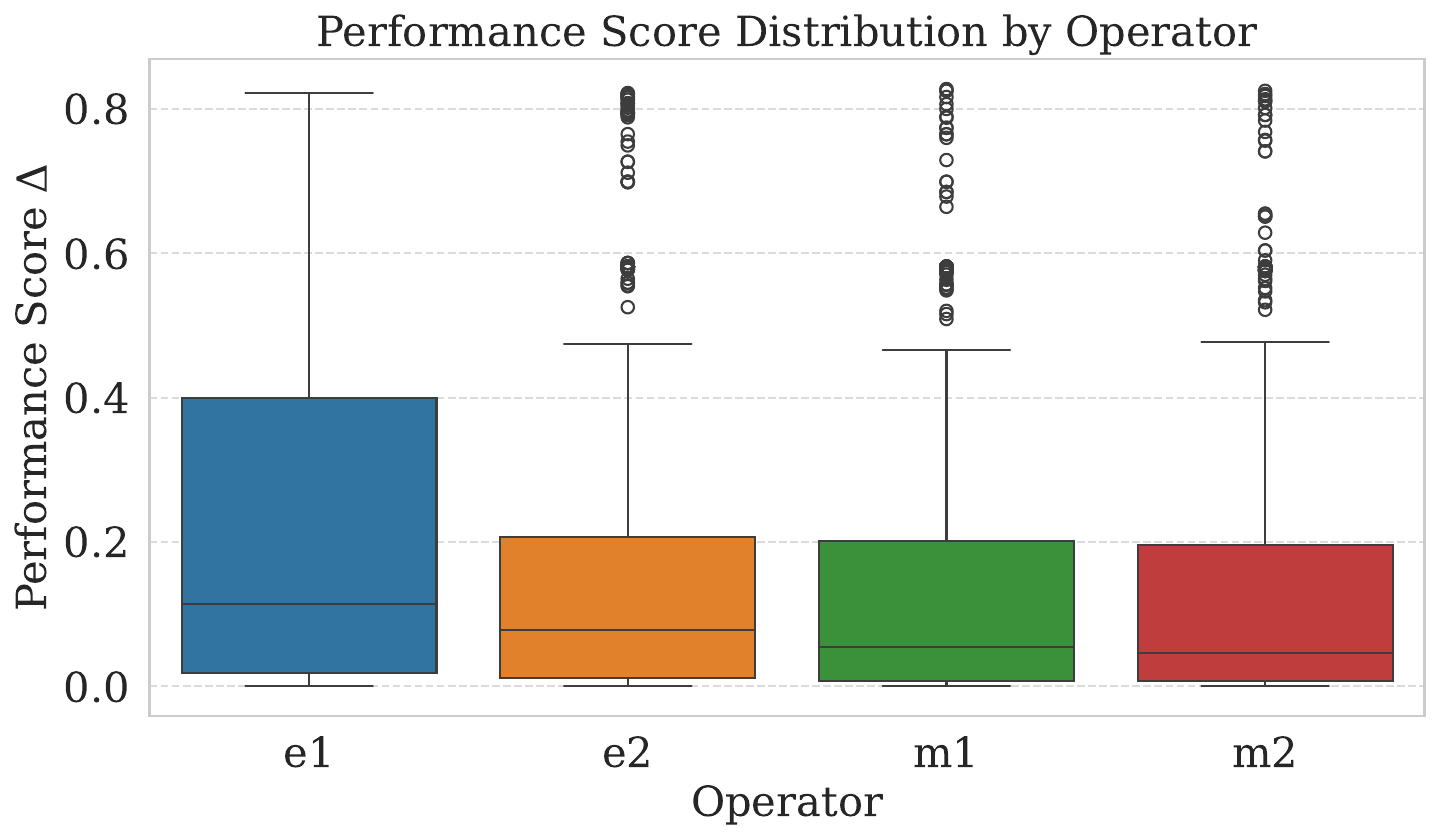}
    \caption{Distribution of performance delta by operator type.}
    \label{fig:operator_perf}
\end{figure}

\begin{figure}[tbp]
    \centering
    \includegraphics[width=0.45\textwidth]{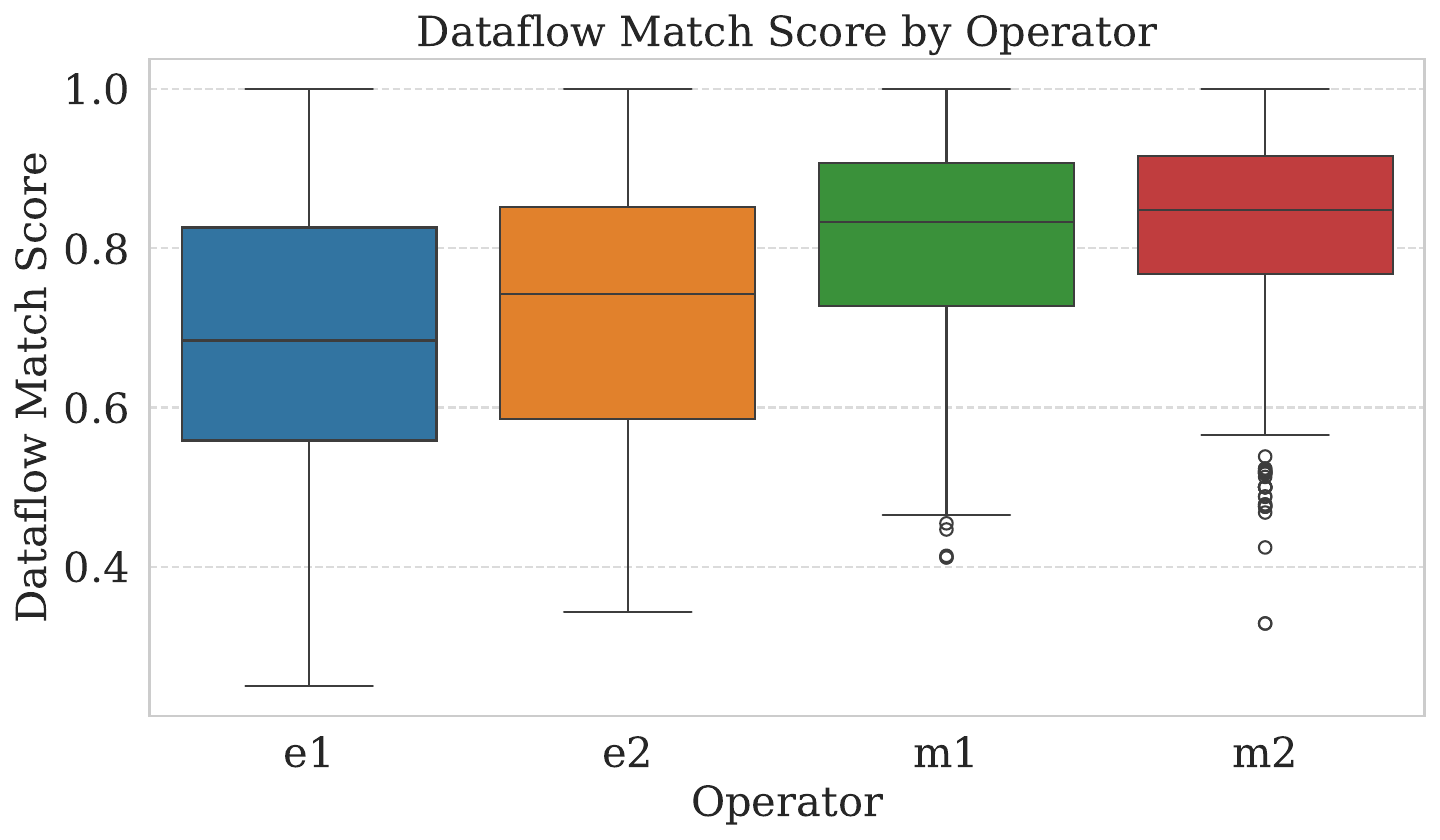}
    \caption{Dataflow Match Score distribution across operator types.}
    \label{fig:operator_dfm}
\end{figure}

\begin{figure}[tbp]
    \centering
    \includegraphics[width=1\linewidth]{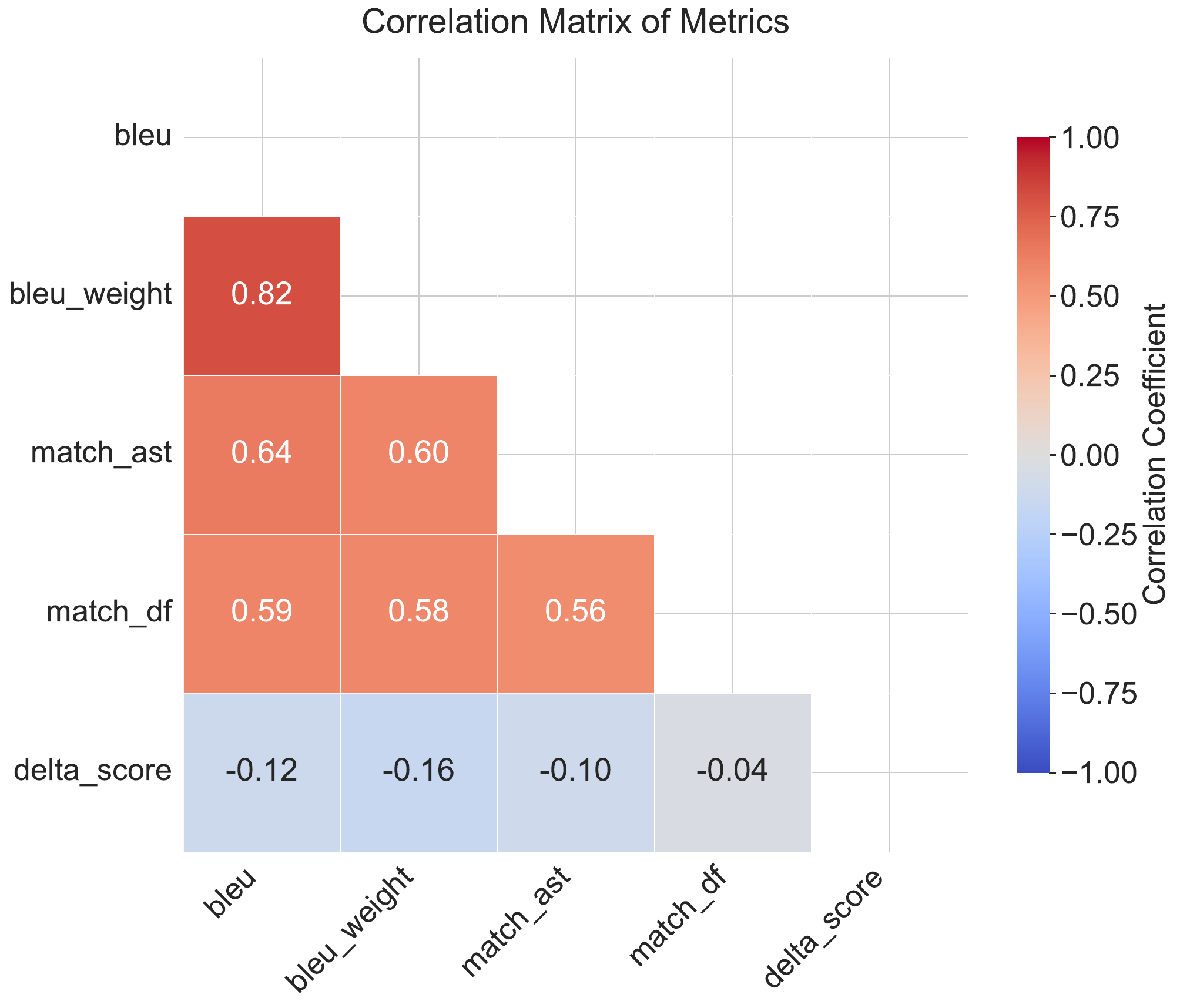}
    \caption{Correlation matrix between four similarity criteria and the performance delta on TSP.}
    \label{fig:enter-label}
\end{figure}

\paragraph{Operator-level Trends.}
To further dissect similarity metrics, we compare distributions across four search operators (E1, E2, M1, M2 used on EoH). These operators are designed for distinct search behaviors: E1 and E2 focus on exploring new algorithms, while M1 and M2 refine existing parent algorithms through exploitation. Fig.~\ref{fig:operator_dfm} and Fig.~\ref{fig:operator_perf} show the performance delta scores and dataflow match scores for algorithm pairs generated by each operator.

Fig.~\ref{fig:operator_perf} shows that E1 results in the widest variance and highest median performance delta, indicating a more aggressive impact on algorithm behavior. This aligns with Fig.~\ref{fig:operator_dfm}, where E1 yields the lowest Dataflow Match scores. In contrast, M1 and M2 maintain higher Dataflow Match scores and tighter performance distributions, implying that mutations they perform are more semantically preserving. The results also validate the necessity of using different operators for balancing exploration and exploitation in LLM-driven automated algorithm search.


\paragraph{Limitations.}
Despite the observed correlations, our analysis reveals notable variability across metrics and tasks, with small correlation coefficients to performance delta. This inconsistency highlights the inherent limitations of code-based similarity measurements in capturing algorithmic similarity. Actually, even the same algorithm can have many different code implementations. These limitations highlight a critical research gap, i.e., current code similarity metrics, while informative, inadequately capture the complex relationship between algorithmic modifications and performances in LLM-assisted search. Future work should focus on developing specialized similarity measurements that better account for algorithmic semantics rather than surface-level syntax, potentially incorporating execution traces, algorithmic complexity analysis, or learned representations that capture functional similarity. 


\section{Conclusion}\label{sec:conclusion}

In this paper, we study the fitness landscape of LLM-assisted automated Algorithm Search (LAS) using a graph-based approach. We try to advance our understanding of LLM-assisted algorithm search, offering insights for LAS behaviours and paving the way for more efficient and effective automated algorithm design. Through extensive experiments on six diverse algorithm design tasks and six commonly used LLMs, we uncovered several key insights.
\begin{itemize}
    \item The fitness landscapes of LLM-driven algorithm search are multimodal and rugged, particularly in heuristic design tasks for combinatorial optimization problems. These landscapes exhibit numerous local optima with similar fitness values, making the search process challenging.
    \item  The structural properties of the landscape vary significantly across tasks and LLMs. For instance, heuristic design tasks often concentrate algorithms in low-fitness regions, while symbolic regression tasks show a stark contrast with most algorithms far from optimal. The choice of LLM and search framework settings, such as population size, impacts the exploration-exploitation trade-off, with larger populations enabling broader exploration.
    \item Our evaluation of four code-based similarity metrics for measuring algorithmic distances reveals both promise and limitations: while we observe correlations between code similarities and performance, confirming that different evolutionary operators enable varying degrees of search space exploration. The notable variability observed across these measurements highlights a critical research gap and underscores the need for developing specialized similarity metrics developed specifically for LAS.
\end{itemize}

Future research could explore tailored similarity metrics tailored for algorithm search would enable more accurate landscape characterization. Additionally, investigating the role of different search frameworks in shaping the landscape could reveal effective strategies for navigating the algorithm space. Understanding these aspects could lead to more effective LLM-assisted automated algorithm design systems.


\bibliographystyle{IEEEtran}
\bibliography{reference}

\end{document}